\documentclass[lettersize,journal]{IEEEtran}
\usepackage{amsmath,amsfonts}
\usepackage{algorithmic}
\usepackage{algorithm}
\usepackage{array}
\usepackage{textcomp}
\usepackage{stfloats}
\usepackage{url}
\usepackage{verbatim}
\usepackage{graphicx}
\def\BibTeX{{\rm B\kern-.05em{\sc i\kern-.025em b}\kern-.08em
    T\kern-.1667em\lower.7ex\hbox{E}\kern-.125emX}}

\usepackage{balance}
\usepackage{amsthm,amsmath,amssymb}
\usepackage{mathrsfs}
\usepackage{subfigure}
\usepackage{enumitem}
\usepackage{booktabs} 
\usepackage{multirow}
\usepackage{url}
\usepackage{braket}
\usepackage{cite}
\usepackage{flushend} 
\begin{document}

\title{\huge FHT-Map: Feature-based Hierarchical Topological Map for Relocalization and Path Planning}

\author{Kun Song, Wenhang Liu, Gaoming Chen, Xiang Xu, and Zhenhua Xiong, \IEEEmembership{Member, IEEE}
\thanks{
		This work was in part supported by the National Natural Science Foundation of China (U1813224) and MoE Key Lab of Artificial Intelligence, AI Institute, Shanghai Jiao Tong University, China. \textit{(Corresponding author: Zhenhua Xiong.)}
	}
	\thanks{
		Kun~Song, Wenhang~Liu, Gaoming Chen, Xiang Xu, and Zhenhua~Xiong are with the School of Mechanical Engineering, Shanghai Jiao Tong University, Shanghai, China (e-mail: coldtea@sjtu.edu.cn; liuwenhang@sjtu.edu.cn; cgm1015@sjtu.edu.cn; xu-xiang@sjtu.edu.cn; mexiong@sjtu.edu.cn).
	}
}


\maketitle

\begin{abstract}
Topological maps are favorable for their small storage compared to geometric map. However, they are limited in relocalization and path planning capabilities.
To solve this problem, a feature-based hierarchical topological map (FHT-Map) is proposed along with a real-time map construction algorithm for robot exploration.
Specifically, the FHT-Map utilizes both RGB cameras and LiDAR information and consists of two types of nodes: main node and support node.
Main nodes will store visual information compressed by convolutional neural network and local laser scan data to enhance subsequent relocalization capability.
Support nodes retain a minimal amount of data to ensure storage efficiency while facilitating path planning.
After map construction with robot exploration, the FHT-Map can be used by other robots for relocalization and path planning.
Experiments are conducted in Gazebo simulator, and the results demonstrate that the proposed FHT-Map can effectively improve relocalization and path planning capability compared with other topological maps.
Moreover, experiments on hierarchical architecture are implemented to show the necessity of two types of nodes.
\end{abstract}


\section{Introduction}
Robot exploration aims to autonomously explore unknown environments, while performing Simultaneous Localization and Mapping (SLAM) to construct a map. 
In the past years, most of the work focused on constructing the map as an accurate geometric map, like point cloud map, occupancy grid map, or truncated signed distance function (TSDF) map. 
These maps show satisfying capabilities in relocalization and path planning. 
However, since geometric maps need provide a precise description of obstacles, they are faced with increasing storage demands in large-scale environments\cite{ort2018autonomous,gomez2020hybrid}.

Therefore, the topological map was proposed \cite{kuipers1988navigation}, which generally consists of nodes representing spatial locations and edges representing traversable paths \cite{yang2021graph,cano2022navigating}. 
Correspondingly, it can be described by undirected graph, which can be applicable for robots in a more storage-efficient manner compared to geometric maps \cite{gomez2020hybrid,yang2021graph,niijima2020city,zhang2022mr}.

Typically, there are two main purposes for the topological map utilization by other robots.
The first purpose is relocalize the robot in the environment, which requires abundant features in specific nodes \cite{liu2022360st,wiyatno2022lifelong,he2022online}.
The second is path planning, which requires a full spatial representation of global environment, where features in nodes are less important.
So there is a dilemma under the constrained storage capacity: whether to store more information within each node for better relocalization capability while reducing the number of nodes, or to store less information within each node while creating more nodes for better path planning capability.

\begin{figure}[!t]
	\centering
	\includegraphics[width=3.4in]{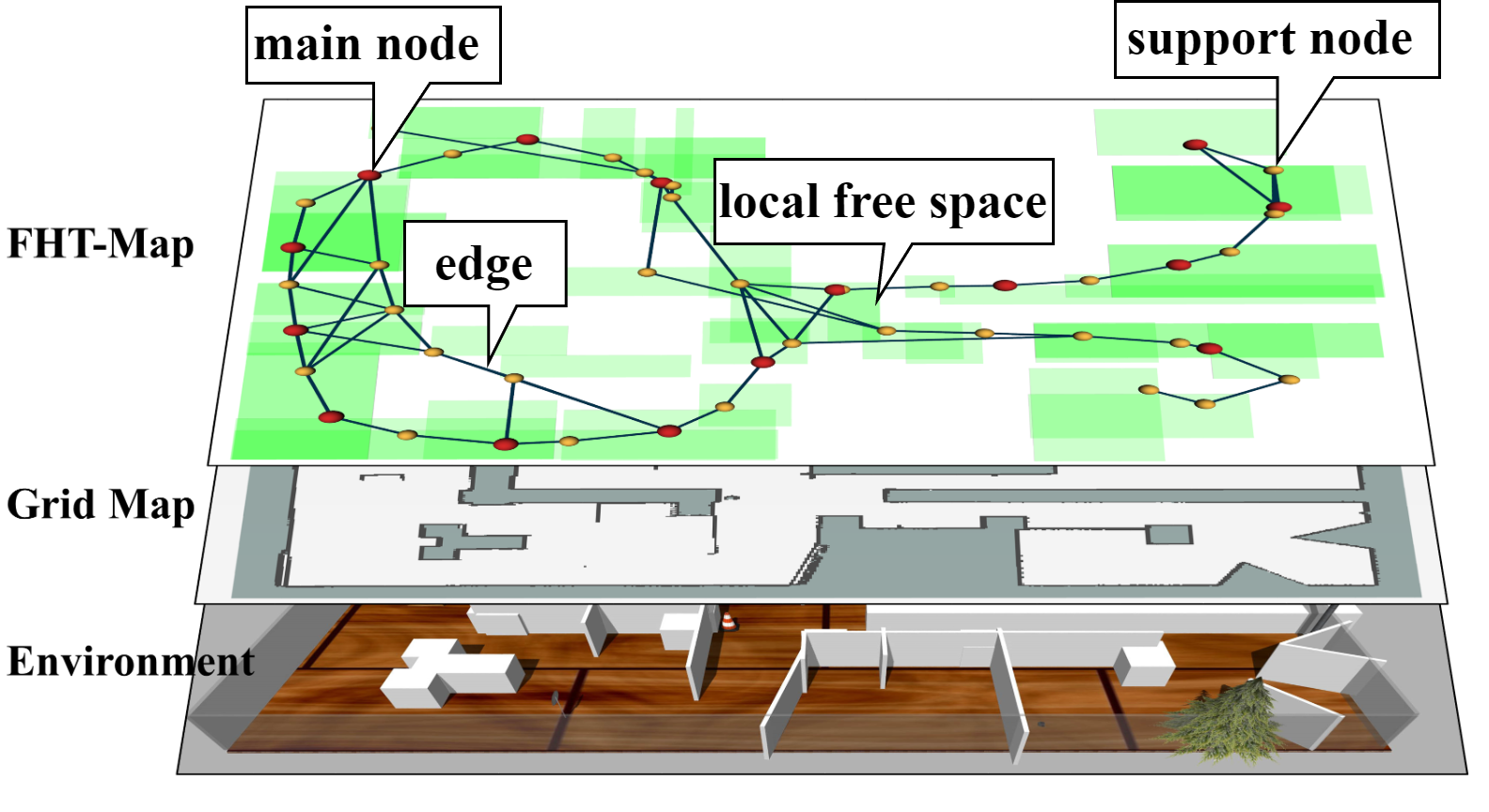}
	\caption{FHT-Map of indoor scene. The red points represent main nodes, the yellow points represent support nodes, and the green rectangles represent the local free space around each node. The edges connecting two nodes represent a traversable path in the environment.}
	\label{topomap}
\end{figure}

Thus, a novel \textbf{F}eature-based \textbf{H}ierarchical \textbf{T}opological \textbf{Map} (FHT-Map) is proposed in this paper. 
There are two types of nodes: main nodes and support nodes, where main nodes store features extracted from the environment and can facilitate relocalization, while minimal amount of data are stored in support nodes, which improves the capability of path planning with very low storage requirements.
Additionally, information about the local free space is also stored in each node. 
Every edge, which connects two nodes, represents a traversable path in the environment. 
Figure \ref{topomap} shows an example of the FHT-Map created for an indoor scene.

In order to achieve real-time construction of the FHT-Map, an autonomous exploration strategy based on a 2D LiDAR is conducted (Section \ref{sec:exp_str}). 
During the exploration process, main nodes are generated at locations with rich visual information in the environment, and support nodes are generated to ensure traversability and path refinement (Section \ref{sec:map_building}).
Additionally, algorithms for construction of edges and local free spaces of nodes are also proposed (Section \ref{sec:map_building}). 
Finally, relocalization and path planning algorithms are proposed (Section \ref{sec:roloca_nav}). The main contributions are:

\begin{itemize}[leftmargin=*]
	\item A novel FHT-Map is proposed, enabling efficient and flexible environment representation with a reduced data volume.
	\item Real-time FHT-Map construction can be realized during autonomous exploration, alongwith the node selection and map refinement algorithms.
	\item Relocalization and path planning algorithms based on FHT-Map are proposed for effective map utilization.
\end{itemize}

\section{RELATED WORK}
\begin{figure*}[!t]
	\centering
	\includegraphics[height=1.9in]{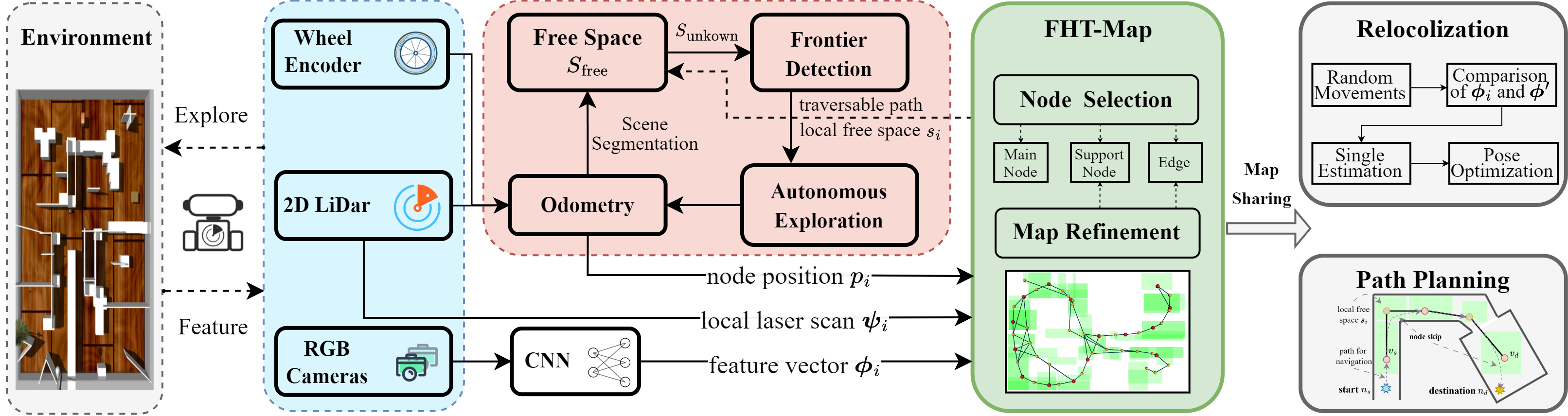}
	\caption{The framework for FHT-Map construction and utilization. The framework first implements a 2D LiDAR based robot exploration strategy. And during this process, FHT-Map is constructed with node selection and map refinement algorithms. After that, the FHT-Map is transmitted to other robots, and with the implementation of relocalization algorithm, robots can use the FHT-Map for path planning and other tasks.}
	\label{fig:total}
\end{figure*}

Topological maps have been widely applied in representation of large-scale environments \cite{gomez2020hybrid,islam2020framework,yang2021graph,niijima2020city} and communication of multi-robot systems\cite{zhang2022mr,patil2023graph,bayer2021decentralized}, due to their efficiency in information storage. In general, topological maps can be categorized by whether it is feature-based or not, where features can be considered as data that represents specific environment information near nodes \cite{goedeme2007omnidirectional}.

\subsection{Non-feature-based Topological Map}
Some researchers focused on constructing topological maps of obstacles without features, indicating untraversable spaces in the environment. 
In \cite{zhou2021raptor, gao2023obstacle}, geometric shapes were utilized to build topological maps of obstacles, which can be used to find a high-quality path in limited time for quadrotor. 

Other researchers focused on constructing topological maps in the free space using undirected graph.
In \cite{yang2021graph,cano2022navigating}, topological maps of free spaces were constructed for underground environments, such as caves. Wang \textit{et al.} \cite{wang2019efficient} proposed a semantic road map (SRM) for indoor scenes, considering both navigable regions and semantic information in the environment. 
Musil \textit{et al.} \cite{musil2022spheremap} introduced SphereMap, which utilizes a spherical representation to represent three-dimensional free spaces, thereby enabling path planning for Unmanned Aerial Vehicles (UAV) in underground environments. 

Since this type of map lacks features, as a result, robots lack the capability to accurately identify and locate themselves in the environment, leading to difficulties in relocalization.
Unreliable relocalization may cause deviations or errors in following the planned path, which even can make the well-planned paths meaningless.

\subsection{Feature-based Topological Map}

To solve this problem, the second category of research focuses on constructing topological maps with features in nodes.
Liu \textit{et al.} \cite{liu2022360st} utilized panoramic camera to build topological maps and nodes were constructed at the location of keyframes. Wiyatno \cite{wiyatno2022lifelong} utilized an RGB-D camera to build the topological map and relocalization is achieved using RGB-D images. 
In \cite{he2022online}, point clouds are stored within the nodes. 
Nonetheless, storing raw features of environment, like images and point clouds, will result in increased storage requirements. 
Zhang \textit{et al.} \cite{zhang2022mr} employed a neural network to extract feature vectors of raw images, enabling rough relocalization with lower storage demand. 

These topological maps store different information in nodes and can be used for relocalization.
However, they lack descriptions of local free spaces around nodes, leading to suboptimal path planning performance for robots.

Therefore, hybrid topological maps were proposed with local free spaces information stored in each nodes.
In \cite{gomez2020hybrid,blochliger2018topomap,he2021hierarchical}, multi-layered topological maps were proposed. 
Free spaces, extracted from point clouds of the environment, were stored in each node. 
These free spaces were connected by edges and capability of path planning can be refined.

Theoretically, the above topological maps should be capable of fulfilling all functions.
However, under limited storage capacity, storing features in every node will lead to a reduced number of nodes and compromise path planning capability.
Hence, there is always a trade-off between relocalization and path planning capabilities.

In this paper, we aim to enhance relocalization and path planning capabilities of topological map while maintaining low storage requirements.
Thus, a novel FHT-Map is proposed.
Main nodes, in which compressed environment information is stored, can achieve relocalization while maintaining the advantage of small storage. 
Lightweight support nodes are generated to represent navigable path in the environment.
Furthermore, a map construction algorithm is presented to control the density of nodes, enabling flexible adjustment of relocalization capabilities and storage volume. 
Finally, relocalization and path planing algorithms are proposed for FHT-Map utilization.

\section{METHODOLOGY}

The framework of the FHT-Map construction and utilization is shown in Figure \ref{fig:total}, which keeps data of three types of sensors as inputs, including wheel encoders, 2D LiDAR, and RGB cameras. Firstly, a robot can employ LiDAR-based autonomous exploration similar to \cite{umari2017autonomous}. During exploration, the map construction algorithm, including node selection and map refinement, is utilized to build the FHT-Map. Then, the FHT-Map can be transmitted to other robots, where relocalization and path planning capabilities can be realized together.

In this section, the detailed processes for environment exploration, FHT-Map construction, and FHT-Map utilization will be introduced.

\subsection{Environment Exploration}
\label{sec:exp_str}
Different kinds of sensors can be used for environment exploration, like camera \cite{bonetto2022irotate} and LiDAR \cite{umari2017autonomous}.
In general, during exploration, the environment can be divided into three parts, free space $S_\text{free}$, occupied space $S_\text{occ}$ and unknown space $S_\text{unknown}$. 
An efficient strategy to expand the explored area is moving the robot towards frontiers, which are defined as the boundaries between $S_\text{free}$ and $S_\text{unknown}$ \cite{yamauchi1997frontier}. 

In this work, the FHT-Map construction process relies solely on estimation of the robot's position and the segmentation of $S_\text{free}$ in the environment. 
It is not sensitive to the specific sensor types used for exploration.
For the sake of simplicity, Cartographer \cite{hess2016real}, which is based on lightweight 2D LiDAR, is used for odometry as well as the segmentation of $S_\text{free}$. 

To achieve autonomous exploration, target of the robot is selected among these frontiers $\{f_i\}$ based on the concept of Next Best View (NBV) \cite{connolly1985determination}. 
Similar to \cite{umari2017autonomous}, a cost-utility function $U(f_i)$ which considers the information gain and navigation cost of frontiers is proposed to find NBV:
\begin{equation}
	\label{equ:U_f}
	U(f_i) = \text{Infor}(f_i) \cdot \text{Cost}^{-1}(f_i,\zeta(t_\text{now}))
\end{equation}
where $\zeta(t_\text{now})$ is the current robot position during exploration. Frontier with the maximum value of $U(f_i)$ will be selected as the next exploration target.

Based on the above strategy, the robot can explore the environment autonomously, thereby ensuring the subsequent construction process of FHT-Map.

\subsection{FHT-Map Construction}
\label{sec:map_building}
In the process of exploration, a sequence of positions $\{\zeta(t)\}$ are determined and FHT-Map can be built during this process. 

FHT-Map can be represented using an undirected graph $\mathcal{G} = (\mathcal{V},\mathcal{E})$, where $\mathcal{V}$ represents the set of nodes and $\mathcal{E}$ represents the set of edges. There are two kinds of nodes in the graph, $\{v_\text{main}\} \subset \mathcal{V}$ is called main node and $\{v_\text{sup}\} \subset \mathcal{V}$ is called support node. Each edge $e = (v^m,v^n)\in \mathcal{E}$ connects two nodes with id $m$ and $n$, and $e \in S_\text{free}$, representing a traversable path between the two nodes in free space.

Contents of main node and support node are different. Main node contains all information presented in Table \ref{table:node_info}, and support node only contains id $i$, position $p_i$ and rectangular free space $s_i$, which is the hierarchical architecture of the FHT-Map.

\begin{table}[htp]
	\renewcommand{\arraystretch}{1.4}
	\caption{Contents Stored in a Node}
	\label{table:node_info}
	\centering
	\begin{tabular}{|c||c|}
		\hline
		$i$ & id of the node in $V$\\
		\hline
		$p_i$ & position of node $i$ in map frame\\
		\hline
		$s_{i}$ & rectangular local free space of node $i$\\
		\hline
		$\boldsymbol{\phi}_i$ & feature vector extracted from RGB images at $p_i$\\
		\hline
		$\boldsymbol{\psi}_i$ & local laser scan at $p_i$\\
		\hline
	\end{tabular}
\end{table}

\subsubsection{Establishing  Main Node}
In order to enable relocalization capability of the FHT-Map, compressed local environmental features are stored in main nodes. 

Due to potential repetition of point cloud features in large-scale environments, we utilize compressed visual features $\{\boldsymbol{\phi_i}\}$ from images to achieve rough relocalization, and local laser scans $\{\boldsymbol{\psi_i}\}$ from 2D LiDAR are used for accurate result. 

Initially, the robot is equipped with a multi-camera system that captures images from 360° perspective. This ensures comprehensive information acquisition from all directions.

Then a convolutional neural network (CNN) \cite{radenovic2018fine} for image retrieval is employed to extract a unit feature vector $\boldsymbol{\phi_i}\in \mathbb{R}^{512}$ from the images. This vector is solely determined by its position in the environment and is orientation-independent, so it can be used as a place recognition (PR) descriptor \cite{dong2022mr}.

For any given time $t$ during the robot's exploration in the environment, assuming its position is denoted as $\zeta(t)$, a feature vector $\boldsymbol{\phi}(t)$ at $\zeta(t)$ can be extracted. It is highly redundant to compute and store $\boldsymbol{\phi}(t)$ at every time step. Therefore, an algorithm for main nodes selection is proposed.

The algorithm is based on the following assumption: the richness of visual information varies across different locations in the environment. Furthermore, we observe that, for humans, locations with richer visual information in space are more helpful for relocalization, which is same for robots. 

Thus, information entropy of $\boldsymbol{\phi}(t)$ can be used to quantify the visual information of a location
\begin{equation}
	I(\boldsymbol{\phi}(t)) = -\sum_{i=1}^n \mathcal{P}_{i} \log \mathcal{P}_i
\end{equation}
where the interval $[0, 1]$ is divided into $n$ equal sub-intervals and $\mathcal{P}_i$ represents the probability of each component in $\boldsymbol{\phi_i}$ falling into the $i$-th sub-interval.

Besides, given limited storage requirements, main nodes should be distributed as widely as possible in the environment, which can improve relocalization capability \cite{he2022online}.

An algorithm for main node selection is proposed considering both visual information richness and density of nodes. For the current topological map $\mathcal{G}$ and robot position $\zeta(t)$, relocalization capability at $\zeta(t)$ is defined as
\begin{equation}
	C(\zeta(t)) = \sum_{i} I(\boldsymbol{\phi}_i) \exp(-\frac{||\zeta(t) - p_i||_2^2}{\sigma_c^2})
\end{equation}
where $\sigma_c$ is a hyper-parameter that related to global relocalization capability of the topological map. Adopting a smaller value will result in a topological map with a higher density of main nodes.

Then, a new main node is selected at $\zeta (t')$ using
\begin{equation}
	\begin{aligned}
		t'&=\mathop{\arg\max} \  I(\boldsymbol{\phi}(t))\\
		& s.t. \quad \gamma_2 < C(\zeta(t)) < \gamma_1.
	\end{aligned}
\end{equation}

\begin{figure}[!t]
	\centering  
	\subfigure[]{ 
		\centering    
		\includegraphics[height=1.5in]{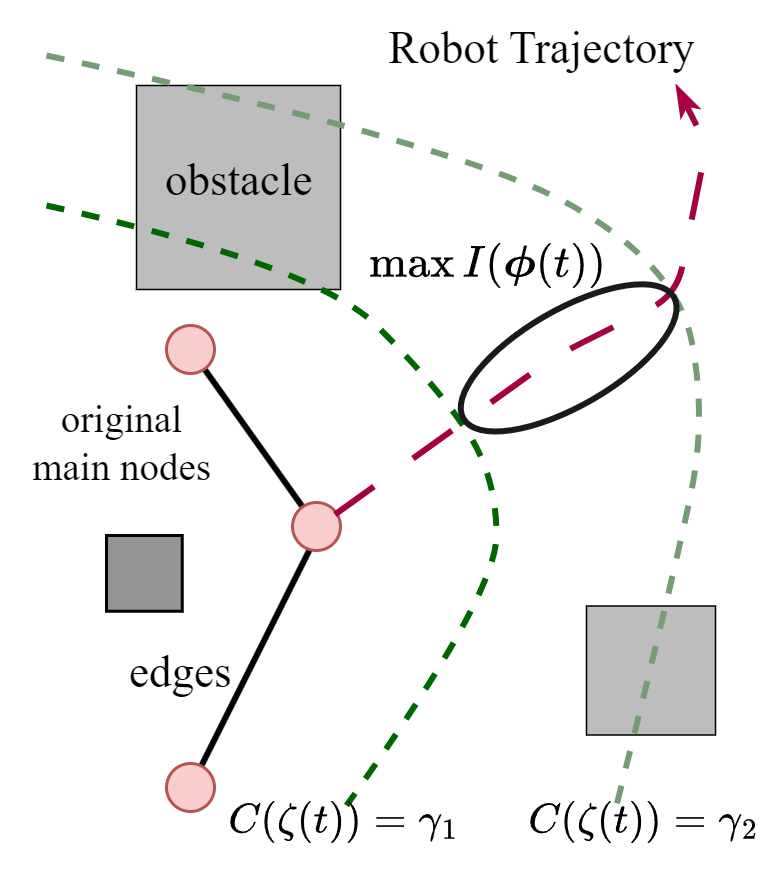}
		\label{fig:main_node}
	}
	\subfigure[]{   
		\centering    
		\includegraphics[height=1.5in]{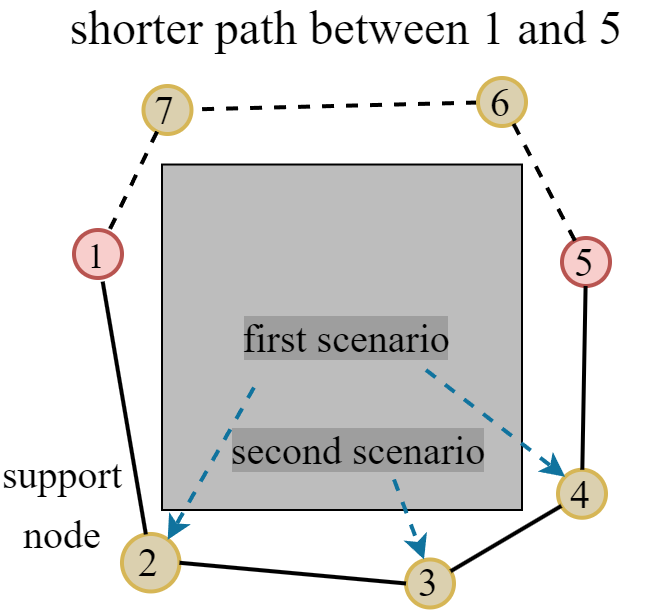}
		\label{fig:sup_node}
	}
	\caption{Illustration of algorithm for creating node. (a) Visualization of process for creating main node.(b) Three situation for creating support node. When the robot moves from node 1 to 5, support nodes 2 and 4 are created to ensure the connectivity of $\mathcal{G}$ (first scenario). Support node 3 is created for second scenario. Additionally, support nodes 6 and 7 are created for map refinement.}    
	\label{fig:node_create}
\end{figure}

As a robot explores the environment, it gradually moves towards unknown areas, leading to a reduction in relocalization capability $C(\zeta(t))$. When it falls below a threshold $\gamma_1$, we consider the robot to enter a region with low relocalization capability. At this point, we start recording a series of potential candidate of main nodes along with their information entropy $I(\boldsymbol{\phi}(t))$. When $C(\zeta(t_\text{now}))$ becomes smaller than  $\gamma_2$, where $\gamma_2 < \gamma_1$, we generate a main node from the candidates with the highest visual information $I(\boldsymbol{\phi}(t))$. The illustration of main nodes construction process is depicted in the Figure \ref{fig:main_node}.

\subsubsection{Establishing Support Node}
Since support nodes only store IDs $i$, positions $p_i$, and local free space $s_i$, their storage demands are smaller compared with main nodes. 
The introduction of support nodes enhances the connectivity of the FHT-Map and can thus improve the capability for path planning, while minimizing storage requirements. 
In this subsection, two scenarios are considered for creating support nodes.

In the first scenario, any potential edge between nodes in $\mathcal{G}$ and current robot position $\zeta(t_\text{now})$ does not belong to the free space $S_{\text{free}}$, which can be described using
\begin{equation}
	\forall v^i \in \mathcal{V}, \ e=(v^i,\zeta(t_\text{now})) \notin S_{\text{free}}.
\end{equation}
In this situation, a support node needs to be created at $\zeta(t_\text{now})$ in order to ensure the connectivity of $\mathcal{G}$.

In the second scenario, a support node is created at $\zeta(t_\text{now})$, when $\zeta(t_\text{now})$ is far from any nodes in $\mathcal{G}$, which means
\begin{equation}
	\forall v^i \in \mathcal{V}, ||v^i - \zeta(t_\text{now})||_2 > th_s
\end{equation}
where $th_s$ is a hyper-parameter to control the density of support node. The introduction of this scenario aims to ensure an adequate density of support nodes, thereby achieving a greater coverage of the local free space $s_i$ in the environment and benefiting path planning.

\subsubsection{Map Update}
After the construction of a node, the algorithms of establishing edge and local free space will be performed for updating the FHT-Map.

When a node $v^i$ is created, the creation of edges is required. An edge should be created between $v^i$ and any other nodes in $\mathcal{V}$ if it satisfies
\begin{equation}
	\forall v^j \in \mathcal{V} \backslash v^i, \ e=(v^i,v^j) \in S_{\text{free}}.
\end{equation}

When $v^i$ is created, we will construct the local free space $s_{i-1} \in S_\text{free}$ for the \textbf{previous} node $v^{i-1}$. The reason for doing this is that when $v_i$ is created, the local free space near $v_i$ is not thoroughly explored, resulting in a limited area of $s_i$. 

Similar to \cite{gao2018online}, rectangular local free spaces in each node can be created by expanding from an initial small rectangle until obstacles in four directions are reached. 

\subsubsection{Topological Map Refinement}
Due to the sensors mounted on the robot, such as the LiDAR, have a certain sensing range, so there are certain navigable paths in the environment that the robot does not need to traverse to complete the exploration. 
Therefore, based on the aforementioned algorithms, the length of planned path will be significantly longer than that based on geometric maps in the worst-case scenario.
One example of this scenario is shown in Figure \ref{fig:sup_node}, where five nodes are created during exploration, but there exists a shorter path between node 5 and 1, leading to a reduced path planning capability between node 1 and 5.

To solve this problem, the algorithm of topological map refinement is proposed.
When a node is constructed in FHT-Map, the distances between this node and other nodes on both the topological map $\mathcal{G}$ and free space $S_\text{free}$ can be obtained using Dijkstra and \textbf{A*} algorithm.
If there exists a node for which the distance on the topological map is significantly longer than the distance on $S_\text{free}$, indicating the existence of a shorter path, a series of sparsely distributed support nodes and edges are created based on the path on $S_\text{free}$.

Since computing the path between two points on $S_\text{free}$ using the \textbf{A*} algorithm is time-consuming, our map refinement algorithm is executed asynchronously with others to ensure real-time construction of FHT-Map.

\subsubsection{FHT-Map Construction}
With algorithms described above, FHT-Map can be constructed during robot exploration, and the overall process is presented in Algorithm \ref{alg:total}.

\begin{algorithm}
	\caption{FHT-Map Construction}
	\label{alg:total}
	\begin{algorithmic}[1]
		\REQUIRE Exploration strategy $\pi$, three different types of sensors $\mathcal{S}$, robot initial position $\zeta(t_0)$, initial free space $S_\text{free}$ 
		\ENSURE FHT-Map $\mathcal{G} = (\mathcal{V},\mathcal{E})$ of the environment
		\STATE $\mathcal{G} \gets \varnothing$, $\zeta(t_\text{now}) \gets \zeta(t_0)$
		\WHILE{\textit{Exploration Not Finish}}
			\STATE $v_\text{main} \gets \texttt{UpdateMainNode}(\mathcal{G},\zeta(t_\text{now}),\mathcal{S})$
			\STATE $v_\text{sup} \gets \texttt{UpdateSupportNode}(\mathcal{G},\zeta(t_\text{now}))$
			\IF{$\{v_\text{main}, v_\text{sup}\} \ne \varnothing$}
				\STATE  $\mathcal{G} \gets \texttt{UpdateMap}(\mathcal{G}, S_\text{free},v_\text{main},  v_\text{sup})$
				\STATE  $\texttt{MapRefinement}(\mathcal{G}, S_\text{free}, v_\text{main}, v_\text{sup})$
			\ENDIF
			\STATE $\zeta(t_\text{now}),\ S_\text{free} \gets \texttt{UpdateRobotPose}(\pi)$
		\ENDWHILE
		\RETURN $\mathcal{G}$
	\end{algorithmic}
\end{algorithm}

The algorithm of FHT-Map construction starts by initializing $\mathcal{G}$ with an empty set. During exploration, nodes and edges are added to $\mathcal{G}$ gradually.
When FHT-Map is constructed, it can be transferred to other robots for utilization.

\subsection{Relocalization and Path Planning}
\label{sec:roloca_nav}
For a constructed FHT-Map, it can be used by other robots \textbf{without knowing} the initial position in the map. To achieve this, a relocalization algorithm is proposed.

Assuming a robot is located in the environment and has obtained FHT-Map, in order to utilize it, the first step is to obtain the transformation matrix $T^\text{map}_\text{odom}\in SE(2)$ between the FHT-Map frame $T_\text{map}$ and the robot odometry frame $T_\text{odom}$.


After relocalization is realized, a path planning algorithm based on FHT-Map is proposed for further utilization.

In this section, we will first present an algorithm for implementing relocalization based on FHT-Map, and then provide a path planning algorithm. 

\subsubsection{FHT-Map based Relocalization}
\label{sec:reloca}

\begin{figure}[!t]
	\centering
	\includegraphics[height=1.9in]{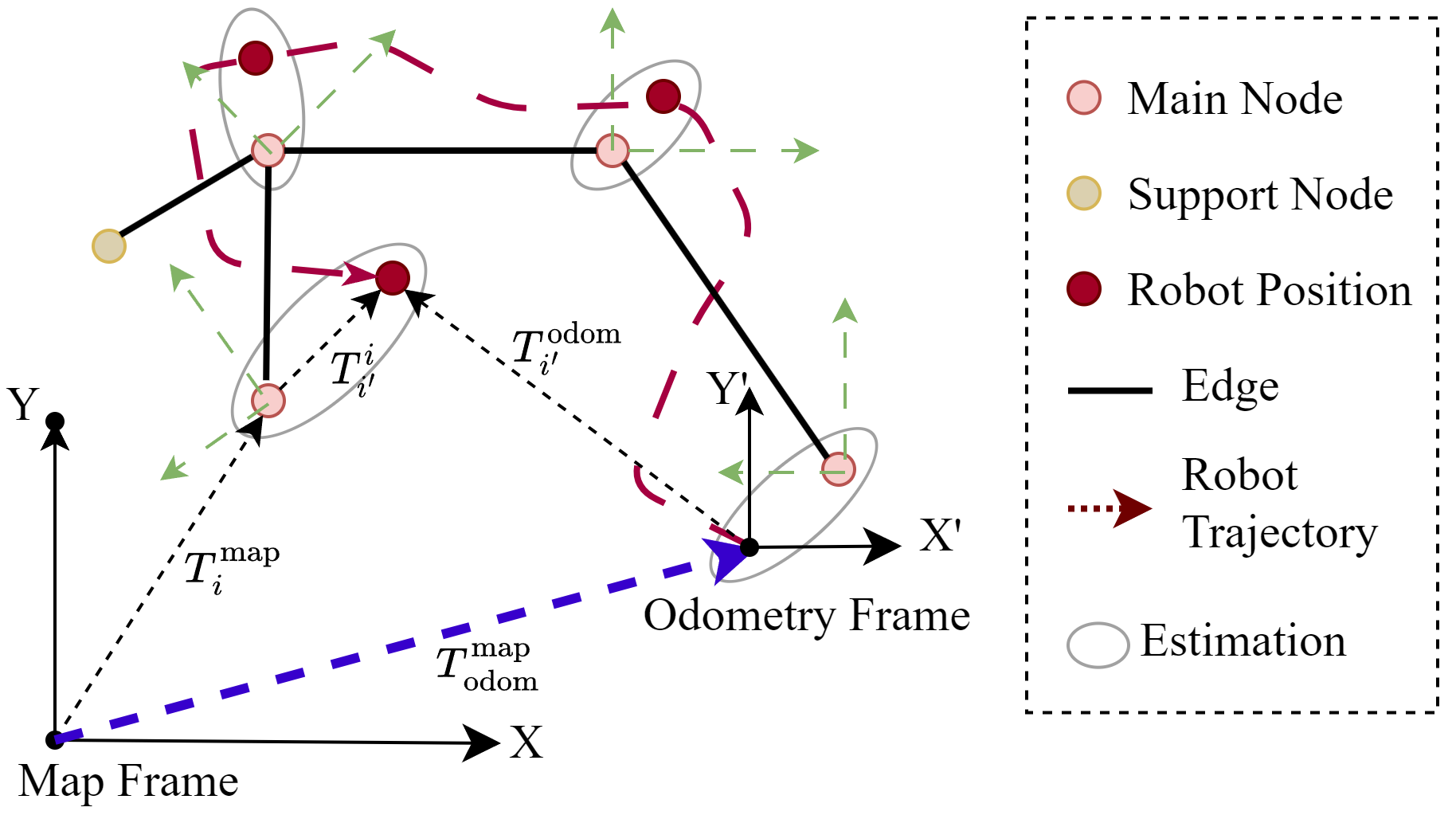}
	\caption{Illustration of robot relocalization. The red dashed line represents the robot's movement trajectory, which passes near four main nodes. Then $T^\text{map}_\text{odom}$ can be calculated by algorithm we proposed, thus achieving relocalization.}
	\label{fig:fht_pose_opt}
\end{figure}
To facilitate relocalization, our approach involves the following steps. After acquiring FHT-Map, the robot initiates random movements within the environment. When robot's current position is in proximity to a main node, a single estimation is obtained. As the robot continues to traverse the environment, multiple estimations are accumulated. Subsequently, a pose graph optimization algorithm is applied to refine relocalization result. The illustration of this process is shown in Figure \ref{fig:fht_pose_opt}.

A single estimation can be obtained through the following algorithm. For the LiDAR and images obtained at the current position of the robot, a visual feature vector $\boldsymbol{\phi}'$ will be extracted firstly. Then inner product between $\boldsymbol{\phi}'$ and feature vectors $\boldsymbol{\phi}_i$ of all main nodes in the FHT-Map will be calculated, the maximum value is
\begin{equation}
	r_{\text{match}} = \mathop{\max}_{i} \boldsymbol{\phi}_i^T \boldsymbol{\phi}'.
\end{equation}

If $r_\text{match}$ is larger than the threshold $th_\text{match}$, then we consider that the robot is near this main node. So we can use the local laser scan $\boldsymbol{\psi}_i$ contained in this main node and the current laser scan $\boldsymbol{\psi}'$ of the robot to calculate the transformation matrix $T^i_{i'}$ between the main node and current position of robot by global Iterative Closest Point (ICP).

\begin{figure}[ht]
	\centering
	\includegraphics[width=3.4in]{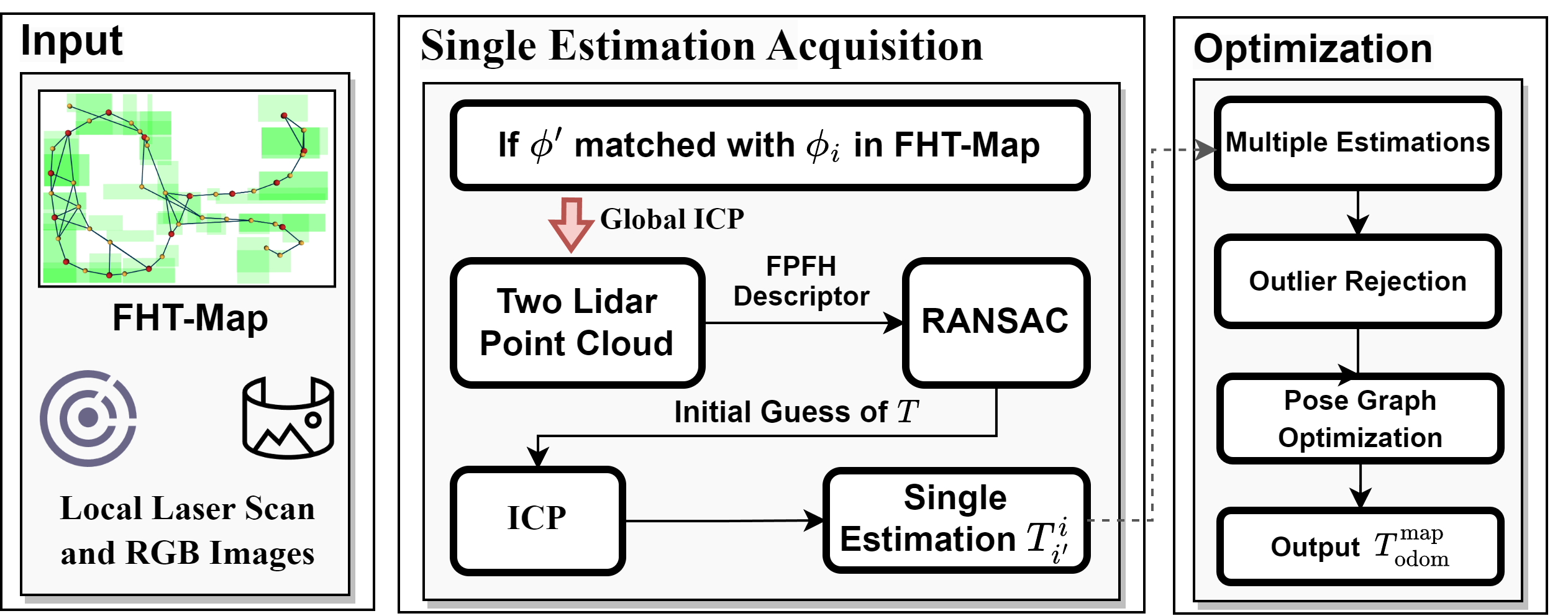}
	\caption{Algorithm for implementing relocalization based on FHT-Map. This algorithm takes FHT-Map and the current laser scan and RGB images of the robot as input, and outputs the transformation matrix $T^\text{map}_\text{odom}$. This algorithm consists of three parts: input, single estimation acquisition, and optimization.}
	\label{fig:reloca_alg}
\end{figure}

Then assuming $n$ estimations $T^i_{i'},\ i=1\cdots n$ are obtained. For the i-th estimation, the transformation of matched node from original map frame is $T^\text{map}_i$, the robot pose of i-th estimation is $T^\text{odom}_{i'}$ and the estimated transformation is $T^{i}_{i'}$. So the i-th estimation of $T^\text{map}_\text{odom}$ can be obtained using
\begin{equation}
	T_{\text{est}_i} = T^\text{map}_i T^{i}_{i'} (T^\text{odom}_{i'})^{-1}.
\end{equation}

For the whole $n$ estimations, we employ an outlier rejection algorithm to remove erroneous estimations firstly. And the final transformation for relocalization can be optimized using 
\begin{equation}
	T_\text{final} = \mathop{\arg\min} \limits_{T^\text{map}_\text{odom}} \sum_i|| T_{\text{est}_i}  \ominus T^\text{map}_\text{odom}||_p
\end{equation}
where $T_\text{final}$ is the optimized result. The algorithm for relocalization based on FHT-Map is illustrated in Figure \ref{fig:reloca_alg}.

\subsubsection{FHT-Map based Path Planning}
After implementing relocalization, we consider using the FHT-Map for path planning. Assume that the robot is currently at a point $n_s$ in the environment and needs to navigate to the destination $n_d$. 

\begin{figure}[t]
	\centering
	\includegraphics[height=1.4in]{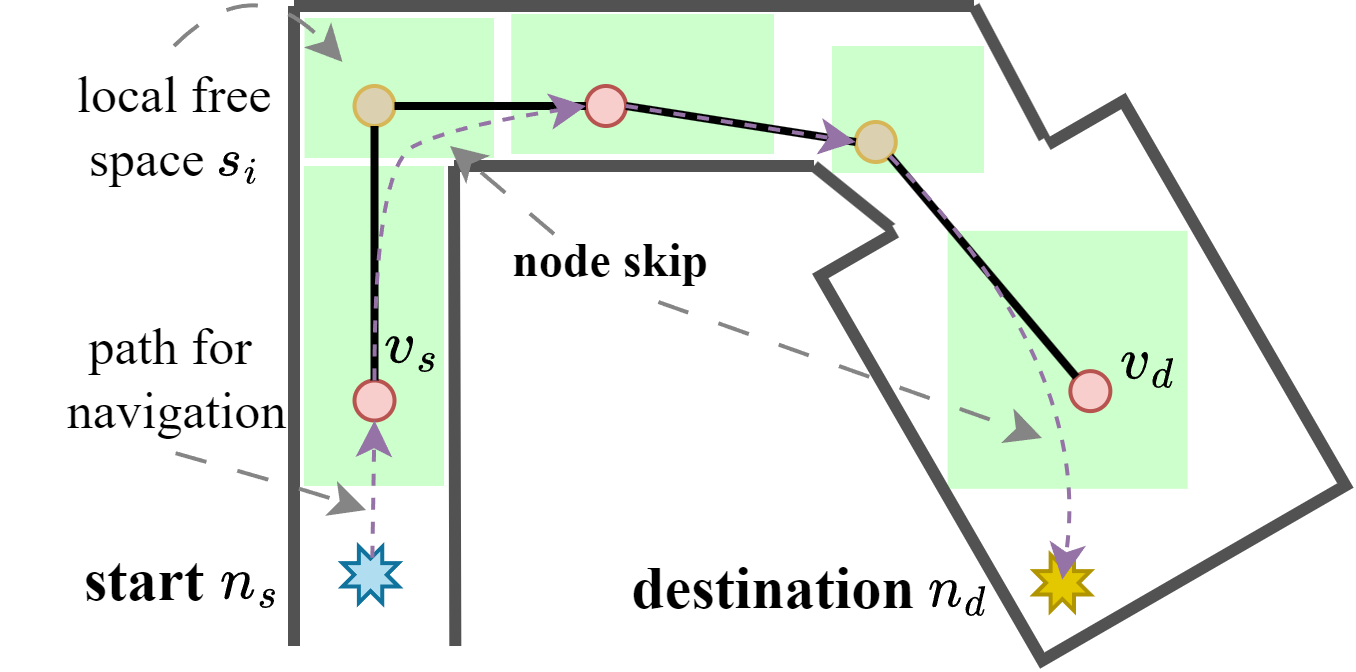}
	\caption{Illustration for robot path planning. The robot first moves to a free space of the nearest node, and then navigates along the topological nodes and edges, finally moving to the destination. During path planning, if the nodes on the subsequent path are already visible, the current target will be skipped, and the robot will proceed directly towards the visible node.}
	\label{fig:nav}
\end{figure}

An overall illustration of robot path planning is shown in Figure \ref{fig:nav}. 
Path planning based on the topological map can be divided into three stages: first, navigate to a specific node in the topological map, then navigate along the topological map, and finally move from the topological map to the destination.

To obtain a path as short as possible, the correct selection of the start node $v^s$ and end node $v^d$ on the topological map is needed.
This problem can be solved using
\begin{equation}
	\label{equ:pathplan}
	\begin{aligned}
	&\min_{v^s,\ v^d \in \mathcal{V}}\  f(n_s, v^s) +  d_{v^s \rightarrow v^d} + f(n_d, v^d)\\
	& s.t.\  
	f(n,p^i)=\left\{
	\begin{aligned}
		||n-p^i||_2,\ &n\in s_i,\\
		k||n-p^i||_2,\  &k\gg1,Otherwise
	\end{aligned}
	\right.
	\end{aligned}
\end{equation}
where $d_{v^s \rightarrow v^d}$ is the shortest distance of $v^s$ and $v^d$ on topological map. 

In Equation (\ref{equ:pathplan}), two situations are considered for $n_s$ and $n_d$. 
If $n_s \in \{s_i\}$, three distance are directly added to determine the shortest path. 
If $n_s \notin \{s_i\}$, the robot will first navigate to the nearest node in $\mathcal{G}$ and then follow the topological map for navigation.
And the same process applies to $n_d$ as well.

Based on the above algorithm, the robot can navigate along $(n_s, v^s , v^d , n_d)$ to achieve FHT-Map based path planning.

\subsubsection{FHT-Map Utilization}
Using the aforementioned algorithms, FHT-Map utilization could be performed.

Assuming the robot is placed in the environment equipped with the same sensors and FHT-Map, the robot first achieves relocalization through random movements. 
Afterward, it can perform path planning. 
During the navigation process, if the robot obtains new estimations, the relocalization results will be further optimized, and path planning will be re-executed to obtain better path.
The overall process for FHT-Map utilization is presented in Algorithm \ref{alg:map_ulti}.

\begin{algorithm}
	\caption{FHT-Map Utilization}
	\label{alg:map_ulti}
	\begin{algorithmic}[1]
		\REQUIRE FHT-Map $\mathcal{G}$, three types of sensors $\mathcal{S}$, random movements trajectory $M$, destination for path planning $n_d$
		\ENSURE Relocalization result $T^\text{map}_\text{odom}$, planned path $p$
		\STATE  $T^\text{map}_\text{odom} \gets \texttt{Relocalization}(\mathcal{G},\mathcal{S},M)$
		\STATE  $p \gets \texttt{PathPlanning}(\mathcal{G},T^\text{map}_\text{odom},n_d)$
		\WHILE{\textit{Destination Not Reached}}
		\IF{\textit{Obtain New Estimation}}
			\STATE $T^\text{map}_\text{odom} \gets \texttt{Optimization}(\mathcal{G},\mathcal{S},T^\text{map}_\text{odom})$
			\STATE $p \gets \texttt{RePlanning}(\mathcal{G},p,T^\text{map}_\text{odom},n_d)$
		\ENDIF
		\STATE $\texttt{navigation}(\mathcal{G},p,n_d)$
		\ENDWHILE
		\RETURN $T^\text{map}_\text{odom},\ p$
	\end{algorithmic}
\end{algorithm}

It's worth noting that if the robot has already obtained an accurate relocalization result, the subsequent path planning process can be performed directly.
Relocalization and path planning are the foundation for map utilization, and our proposed algorithm effectively solves this two problem for FHT-Map, paving the way for more advanced applications.

\section{EXPERIMENT RESULT}

Experiments are conducted in the Gazebo\cite{koenig2004design} simulator. To achieve autonomous exploration, a Turtlebot Burger robot is equipped with a 2D LiDAR sensor (with a range of 7m) and wheel encoders. Additionally, the robot uses four RGB cameras with $90^{\circ}$ Field of View (FOV), and these cameras are arranged horizontally in a circular manner to capture information from all direction.

\begin{figure}[!t]
	\centering  
	\subfigure[]{ 
		\centering    
		\includegraphics[height=0.9in]{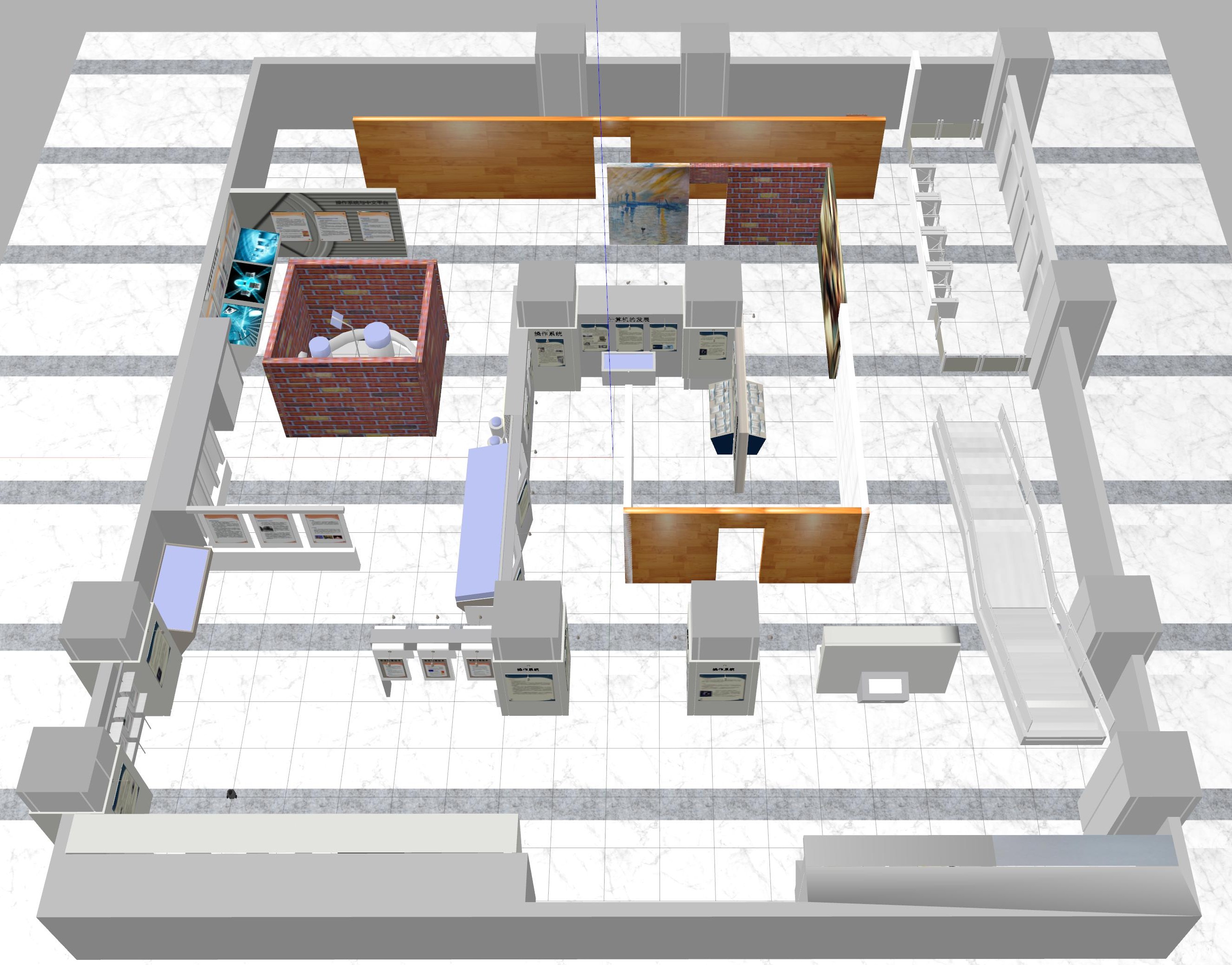}
		\label{fig:museum_env}
	}
	\subfigure[]{   
		\centering    
		\includegraphics[height=0.9in]{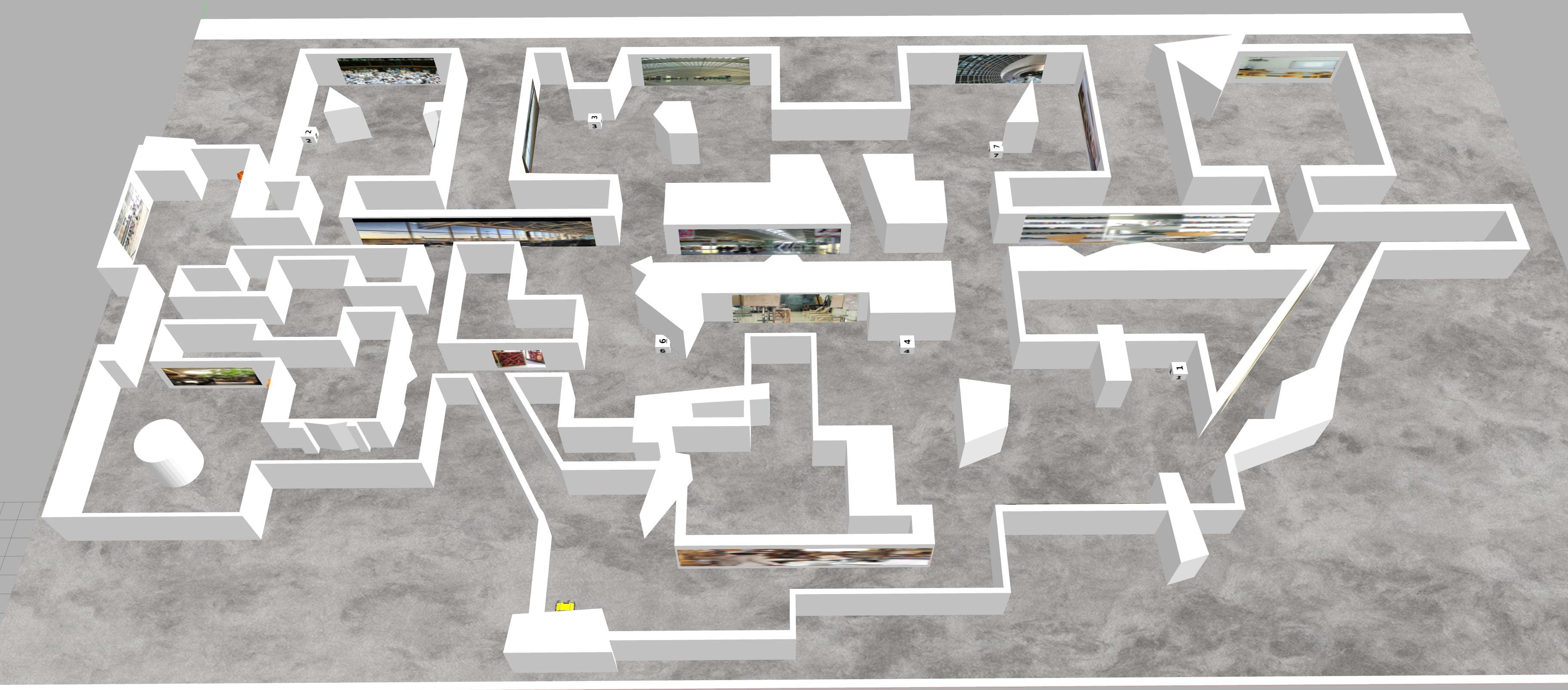}
		\label{fig:office_env}
	}
	\subfigure[]{   
		\centering    
		\includegraphics[height=1in]{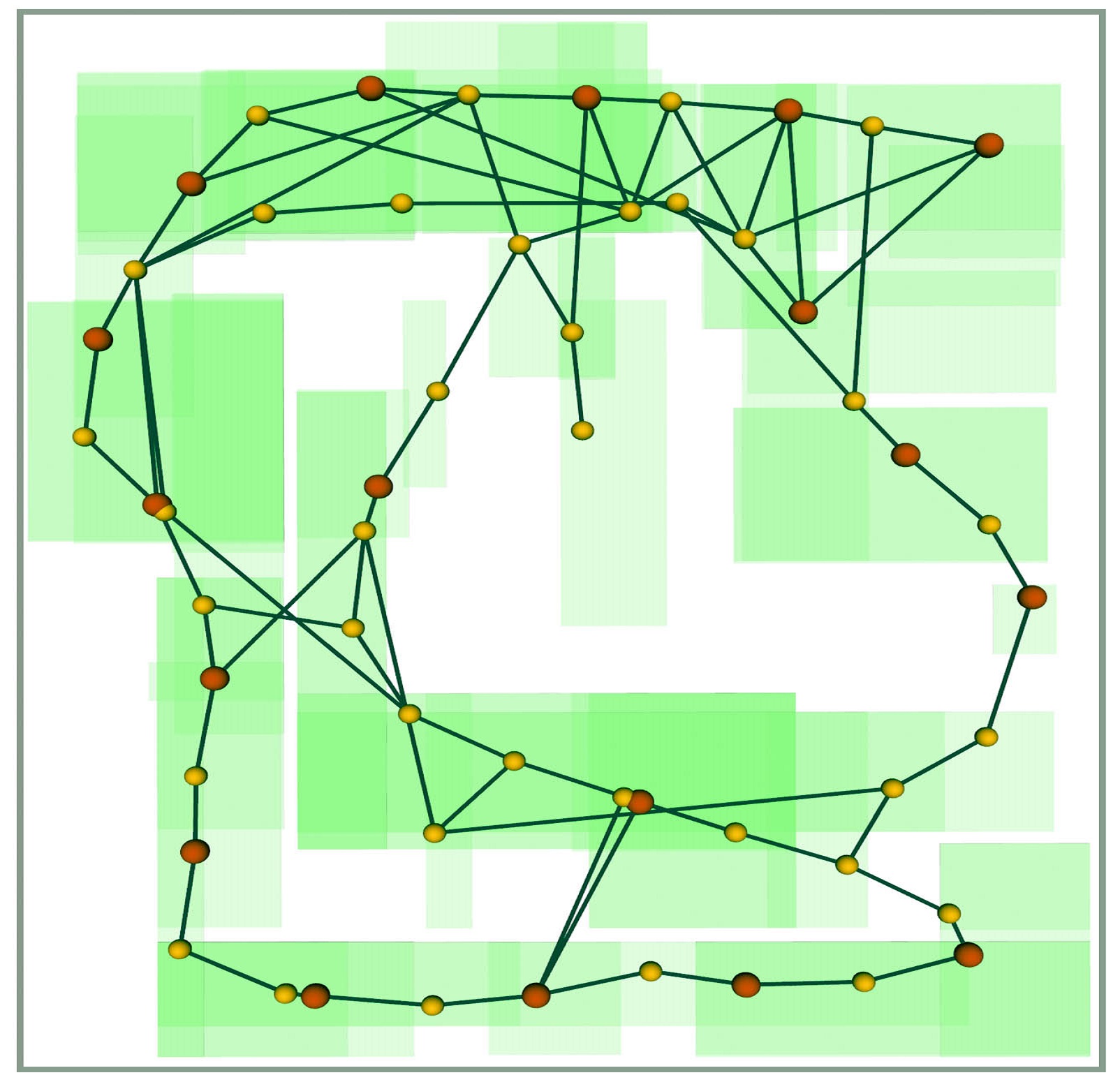}
		\label{fig:museum_res}
	}
	\subfigure[]{   
		\centering    
		\includegraphics[height=1in]{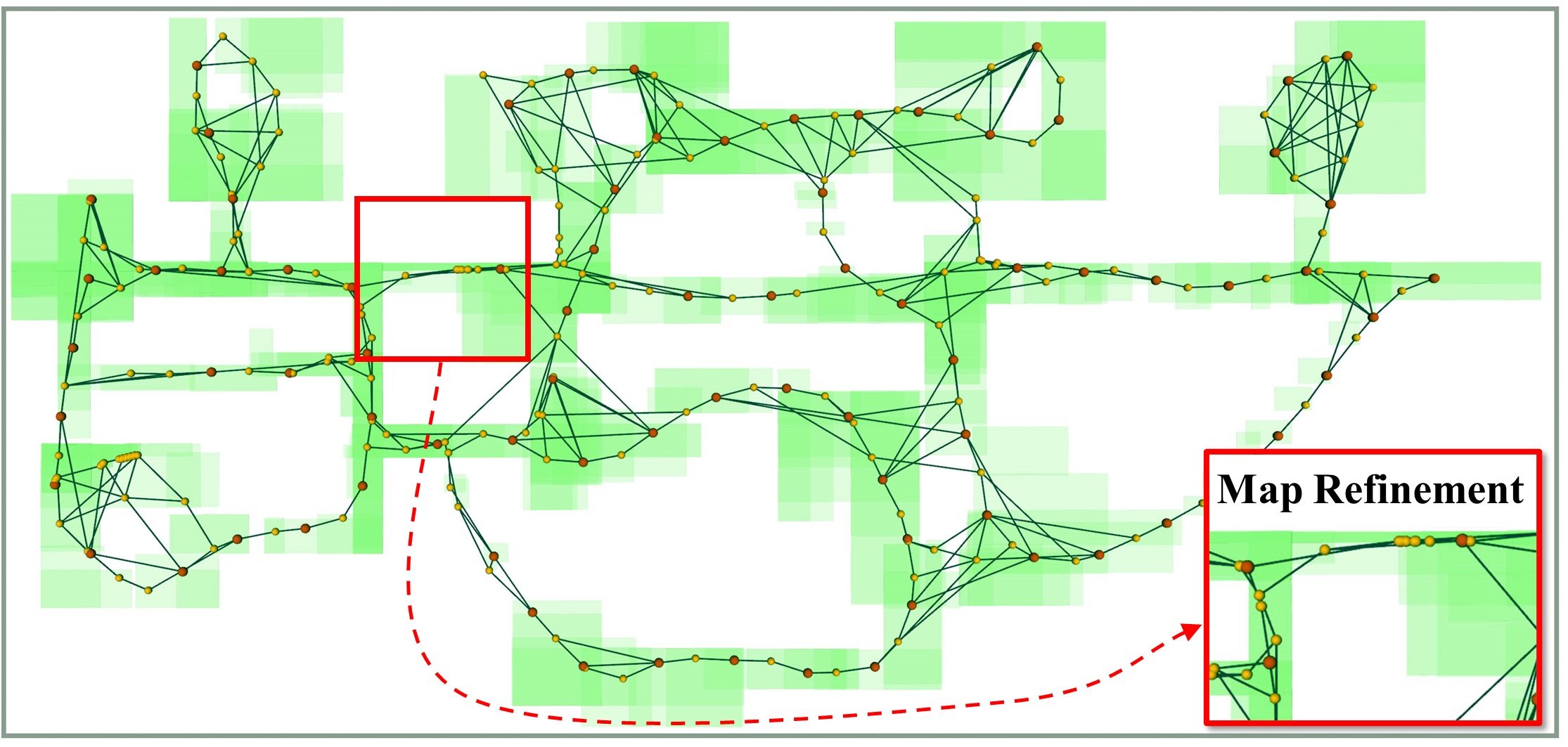}
		\label{fig:office_re}
	}
	\caption{Two simulation environments and their constructed FHT-Maps. (a) Museum environment. (b) Office environment. (c) FHT-Map of Museum. (d) FHT-Map of Office.}    
	\label{fig:two_env}
\end{figure}

Two simulation environments are constructed using Gazebo as depicted in Figure \ref{fig:museum_env} and \ref{fig:office_env}, where the museum environment is 485$m^2$ and the office environment is 3285.22$m^2$. The museum environment, despite its smaller size, is more complex and unstructured. The office environment consists of nine rooms and interconnected corridors, which is simpler and has fewer obstacles.

\subsection{Comparison Experiment with Other Maps}
Storage, relocalization and path planning capabilities of different maps are considered in our work. We compare FHT-Map with a representative approach of feature-based topological map MR-TopoMap \cite{zhang2022mr} and the traditional 2D grid map. 

\subsubsection{Storage Evaluation}
In FHT-Map, the parameter $\sigma_c$ plays a crucial role as it controls the sparsity of the main nodes, which are responsible for storage volume. 
To facilitate a more distinct comparison of relocalization and path planning capabilities, we adopt $\sigma_c = 2.65$ in FHT-Map, aiming for a storage volume that is relatively equivalent to MR-TopoMap.

As shown in Table \ref{table:compare_exp}, compared with grid map, in museum environment, FHT-Map reduces 68.33\% of storage load. And in  office, the storage load is further reduced to 77.56\%.

\subsubsection{Relocalization Evaluation}
In terms of relocalization, for grid map, we employ the widely used Adaptive Monte Carlo Localization (AMCL) algorithm \cite{ros_amcl} as baseline. 
As for the topological map MR-TopoMap, the algorithm in section \ref{sec:reloca} with $T_{i'}^i = I_{4\times4}$ is used for relocalization.

The relocalization algorithm outputs $T^\text{map}_\text{odom} \in SE(2)$, which can be decomposed into translation and rotation components. Assuming the translation component is $t^\text{map}_\text{odom}$ and the rotation component is $\theta^{\text{map}}_\text{odom}$, their ground truth values are denoted as $t_\text{gt}$ and $\theta_\text{gt}$ respectively. 
So the errors for relocalization are

\begin{equation}
	\varepsilon_t = \frac{||t^\text{map}_\text{odom} - t_\text{gt}||_2}{t_\text{gt}},\ \varepsilon_\theta = |\theta^{\text{map}}_\text{odom} - \theta_\text{gt}|.
\end{equation}

Additionally, we focus on the success rate of relocalization, where a successful relocalization is defined as $||t^\text{map}_\text{odom} - t_\text{gt}||_2 < 1$ and $\varepsilon_\theta < 5$ can be reached. 
Besides, the trajectory length $l_\text{reloca}$ of first successful relocalization is counted.

In both two environments, we randomly select eight different initial positions for each. 
Given a motion trajectory, three maps are employed to perform relocalization without an initial guess of $T^\text{map}_\text{odom}$, which means global relocalization is performed. 
We compute mean and standard deviation of four indicators in eight trials, which are shown in Table \ref{table:compare_exp}.

In all experiments, both FHT-Map and MR-TopoMap achieve successful relocalization. 
However, due to its inherent randomness, AMCL has a probability of failure in relocalization. 
In museum, the success rate of AMCL is 85.7\%, while in office, it is only 25\%. 
This decrease in success rate for the office environment can be attributed to its larger size, which presents challenges in relocalization for geometry-based map.

In the museum (office) environment, FHT-Map achieves a relocalization trajectory length $l_\text{reloca}$ that is 40.27\% (35.32\%) of MR-TopoMap's length and 35.11\% (62.16\%) of AMCL's length. 
A shorter trajectory length for relocalization implies faster achievement of relocalization, which is highly desirable for map's applications.

As for relocalization accuracy, in the museum (office) environment, FHT-Map achieves a translation error reduction of 87.00\% (72.49\%) compared to MR-TopoMap and a reduction of 64.64\% (83.38\%) compared to AMCL. 
FHT-Map also exhibits a rotation error reduction of 84.50\% (43.68\%) compared to MR-TopoMap and a reduction of 71.13\% (39.29\%) compared to AMCL.

\begin{table}[!t]
	\fontsize{7}{8}\selectfont
	\renewcommand{\arraystretch}{1.4}
	\caption{Comparison Experiment with Other Maps}
	\label{table:compare_exp}
	\centering
	\begin{tabular}{m{0.6cm}<{\centering}cccc}
		\toprule
		Scene & Indicator & FHT-Map  &  MR-TopoMap & Grid Map\\
		\midrule
		\multirow{6}*{\textbf{Museum}} 
		& Storage\ (KB) & \textbf{60.10} &  64.52 & 189.79\\
		\cmidrule(lr){2-5}
		~& success\ (\%) & 100 & 100 & 87.5\\
		~& $l_\text{reloca}$\ (m)     & \textbf{5.01} $\pm$ 3.55 & 12.44 $\pm$ 6.82 & 14.27 $\pm$ 13.73\\
		~& $\varepsilon_t$\ (\%) & \textbf{0.541} $\pm$ 0.311 & 4.16 $\pm$ 1.85 & 1.53 $\pm$ 0.555\\
		~& $\varepsilon_\theta$\ (degree) & \textbf{0.231} $\pm$ 0.225 & 1.49 $\pm$ 0.64 &0.800 $\pm$ 0.377\\
		\cmidrule(lr){2-5}
		~& $C_\text{path}$ (max) & \textbf{1.08} (\textbf{1.18}) & 1.26 (1.70) & 1 (1)\\
		\cmidrule(lr){1-5}
		\multirow{6}*{\textbf{Office}} 
		& Storage\ (KB) & 278.93 &  \textbf{258.02} & 1243.05\\
		\cmidrule(lr){2-5}
		~& success\ (\%)     & 100 & 100 & 25\\
		~& $l_\text{reloca}$\ (m)     & \textbf{5.47} $\pm$ 4.04 & 16.41 $\pm$ 6.73 & 8.80 $\pm$ 1.93 \\
		~& $\varepsilon_t$\ (\%) & \textbf{0.751} $\pm$ 0.341 & 2.73 $\pm$ 1.85 & 4.52 $\pm$ 3.86\\
		~& $\varepsilon_\theta$\ (degree) & \textbf{0.428} $\pm$ 0.214 & 0.760 $\pm$ 0.853 & 0.705 $\pm$ 0.344\\
		\cmidrule(lr){2-5}
		~& $C_\text{path}$ (max) & \textbf{1.07} (\textbf{1.24}) & 4.76 (14.56) & 1(1)\\
		\bottomrule
	\end{tabular}
\end{table}

\subsubsection{Path Planning Evaluation}
Assuming that a robot needs to move from its current position to another point in the map, the actual path length obtained under topological map is $s_\text{topo}$, and the length on the grid map is $s_\text{grid}$. 
Considering that the grid map has a complete representation of environmental obstacles, $s_\text{grid}$ will be shorter than $s_\text{topo}$. Therefore, we define the capability for path planning is
\begin{equation}
	C_\text{path} = \frac{s_\text{topo}}{s_\text{grid}} >1
\end{equation}
where smaller $C_\text{path}$ indicates better path planning capability.

We randomly select six pairs of different start points and destinations in two environments and compute mean and max value of $C_\text{path}$. The results are shown in Table \ref{table:compare_exp}.

It can be observed that in  museum environment, FHT-Map exhibits a 18\% improvement in path planning capability $C_\text{path}$ over MR-TopoMap in terms of the average value. 
And in extreme cases (maximum value), it shows a 62\% improvement. 
In the larger office environment, FHT-Map improves 3.69 times in path planning capability compared to MR-TopoMap in terms of the average value, and in extreme cases (maximum value), it improves 13.32 times. 
This can be attributed to that all traversable paths in the environment are represented in FHT-Map compared to MR-TopoMap.

\subsection{Experiment on Hierarchical Architecture}
More experiments are conducted to validate the effectiveness of hierarchical architecture. 
We compare two framework with different components: the first framework only has the algorithm for main node selection, referred to as ``Main Only"; and the second framework is the complete version of FHT-Map, which includes the hierarchical architecture along with the map refinement module. 
The capabilities for relocalization and path planning are evaluated in both two environments, with $\sigma_c = 2.65$ for museum and $\sigma_c = 5$ for office.

The results are shown in Table \ref{table:Ablation_exp}. Compared with ``Main Only", the introduction of the hierarchical architecture in FHT-Map reduces the storage requirements by 36.55\% and 28.65\% in the museum and office environments, respectively. 

Although ``Main Only" has more main nodes, its relocalization capability is not improved compared to FHT-Map. 
Additionally, the inclusion of support nodes and map refinement module in FHT-Map contributes to an improved path planning capability $C_\text{path}$. 
In office, $C_\text{path}$ of FHT-Map is 2.62 times better on average compared to ``Main Only" and 7.52 times better in extreme cases (maximum value).

\begin{table}[!t]
	\fontsize{6.5}{8}\selectfont
	\renewcommand{\arraystretch}{1.4}
	\caption{Experiment on Hierarchical Architecture}
	\label{table:Ablation_exp}
	\centering
	\begin{tabular}{m{1.2cm}<{\centering}cccc}
		\toprule
		Scene & \multicolumn{2}{c}{\textbf{Museum}} & \multicolumn{2}{c}{\textbf{Office}}\\
		\cmidrule(lr){2-3}\cmidrule(lr){4-5}
		&  Main Only &  FHT-Map  &  Main Only &  FHT-Map\\
		\midrule
		\tiny{Storage\ (KB)} & 94.72  & \textbf{60.10}& 154.88& \textbf{110.51}\\
		\cmidrule(lr){1-5}
		$l_\text{reloca}$\ (m)     & 5.91 $\pm$ 4.42 & {5.01} $\pm$ 3.55 & {14.29} $\pm$ 8.58 & 14.60 $\pm$ 16.60 \\
		$\varepsilon_t$\ (\%) & {0.511} $\pm$ 0.362 &  0.541 $\pm$ 0.311 & 0.678 $\pm$ 0.621 & {0.665} $\pm$ 0.220\\
		$\varepsilon_\theta$\ (degree) & 0.30 $\pm$ 0.17 & {0.23} $\pm$ 0.22 & {0.439}$\pm$ 0.191 & 0.634 $\pm$ 0.707\\
		\cmidrule(lr){1-5}
		$C_\text{path}$ (max) & 1.18 (1.48)  & \textbf{1.08} (\textbf{1.18}) & 2.72 (8.42) & \textbf{1.04} (\textbf{1.12})\\
		\bottomrule
	\end{tabular}
\end{table}

\subsection{Discussion on Proposed Method}
In general, the reasons for advantages of FHT-Map can be summarized as follows. 
Firstly, the introduction of hierarchical framework in FHT-Map, combining main nodes and support nodes, reduces storage requirements compared with other topological maps and grid maps.

Besides, both visual feature vectors and local laser scans are stored in main nodes. Using main nodes and pose graph optimization algorithm, accurate and fast relocalization can be obtained even in the presence of noise in estimations. Algorithm for main nodes selection, where visual information richness is considered, optimizes the distribution of main nodes, thereby reducing the path length for relocalization.

Additionally, the inclusion of support nodes and map refinement module ensures comprehensive representation of different paths in the environment, which improves the path planning capability in FHT-Map.

Since $\sigma_c$ is an important parameter in this work, its impact on FHT-Map is discussed. A larger $\sigma_c$ directly affects the number of main nodes in FHT-Map, resulting in reduced storage requirement. This, in turn, leads to slower relocalization realization, indicated by an increment in $l_\text{reloca}$. However, after obtaining a sufficient number of observations, it can achieve same relocalization accuracy. As the path planning capability primarily comes from the introduction of support nodes, $\sigma_c$ has minimal impact on it.

\section{CONCLUSION \& FUTURE WORK}
This study presented a framework called FHT-Map and its construction algorithms for the main nodes, support nodes, edges and local free spaces. 
Furthermore, we took into account the requirements for map utilization and developed relocalization and path planning algorithms based on FHT-Map.
Detailed experiments were conducted to validate the effectiveness of the algorithms, demonstrating that FHT-Map achieved improvements in storage demands, relocalization, and path planning capabilities compared to other topological and geometric maps.
The main drawback of FHT-Map is the reliance on multiple hyper-parameters, some of which require manual configuration.

Currently, FHT-Map has only been tested in simulation environment. In the future, we aim to establish a platform for testing in real-world large-scale environments to validate the effectiveness of this system.

\bibliographystyle{ieeetr} 
\bibliography{ref}
\end{document}